\documentclass{article}

\usepackage[final, nonatbib]{nips_2018}

\usepackage[numbers]{natbib}
\usepackage[utf8]{inputenc} 
\usepackage[T1]{fontenc}    
\usepackage{hyperref}       
\usepackage{url}            
\usepackage{booktabs}       
\usepackage{amsfonts}       
\usepackage{nicefrac}       
\usepackage{microtype}      
\usepackage{graphicx}
\usepackage[nottoc]{tocbibind}
\usepackage[english]{babel}
\usepackage[]{algorithm2e}
\usepackage{caption} 
\usepackage{array}
\newcolumntype{P}[1]{>{\centering\arraybackslash}p{#1}}
\usepackage{siunitx}

\title{Improving Generalization for Abstract Reasoning Tasks Using Disentangled Feature Representations}

\author{
  Xander Steenbrugge\\
  ML6 \& IDlab, Ghent University - imec\\
  \texttt{xander.steenbrugge@ml6.eu}
  \And
  Sam Leroux, Tim Verbelen, Bart Dhoedt\\
  IDlab, Ghent University - imec \\
  \texttt{firstname.lastname@ugent.be }
}

\begin{document}
\maketitle

\begin{abstract}
  In this work we explore the generalization characteristics of unsupervised representation learning by leveraging disentangled VAE's to learn a useful latent space on a set of relational reasoning problems derived from Raven Progressive Matrices. We show that the latent representations, learned by unsupervised training using the right objective function, significantly outperform the same architectures trained with purely supervised learning, especially when it comes to generalization.
\end{abstract}

\section{Introduction}

\begin{figure}[h!]
  \centering
  \includegraphics[width=5in]{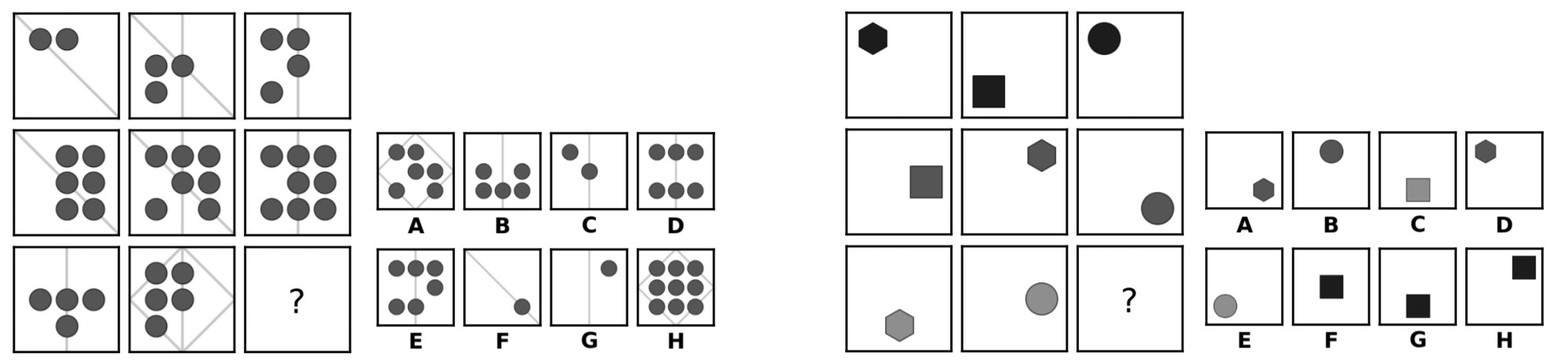}
  \caption{Two example PGM problems: a grid of 3-by-3 related images where the bottom right one is missing and a set of 8 possible answers. The correct choice panels are A and C respectively.}
  \label{fig:RPM_2_examples}
\end{figure}

Reasoning about abstract concepts has been a long standing challenge in machine learning. Recent work by Barret et al.~\cite{WReN} introduces a concrete problem setting for testing generalization in the form of a relational reasoning problem derived from Raven Progressive Matrices that are often used in human IQ-tests. The problem exists of a grid of 3-by-3 related images where the bottom right one is missing and a set of 8 possible answers, of which exactly one is correct. Two examples are shown in Figure \ref{fig:RPM_2_examples}. In this work we use the same dataset, which can be downloaded from \url{https://github.com/deepmind/abstract-reasoning-matrices}.

To create a \textit{Procedurally Generated Matrices} (PGM) dataset, a set of properties is first randomly sampled from the following primitive sets:
\begin{itemize}
  \item Relation types: \ ($R$, with elements $r$): \textit{progression, XOR, OR, AND, consistent union}
  \item Attribute types: ($A$, with elements $a$): \textit{size, type,
colour, position, number}
  \item Object types: \ \ \ ($O$, with elements $o$): \textit{shape, line}
\end{itemize}

The structure \textit{$S$} of a PGM then, is a set of triples, $S = \{[r, o, a]: r\in R,\ o\in O,\ a\in A\}$. These triples determine the challenge posed by one particular matrix problem. In the used dataset, up to 4 triples can be present in a single problem: $1 \leq |S| \leq 4$.\\

\begin{figure}
  \centering
  \includegraphics[width=\textwidth]{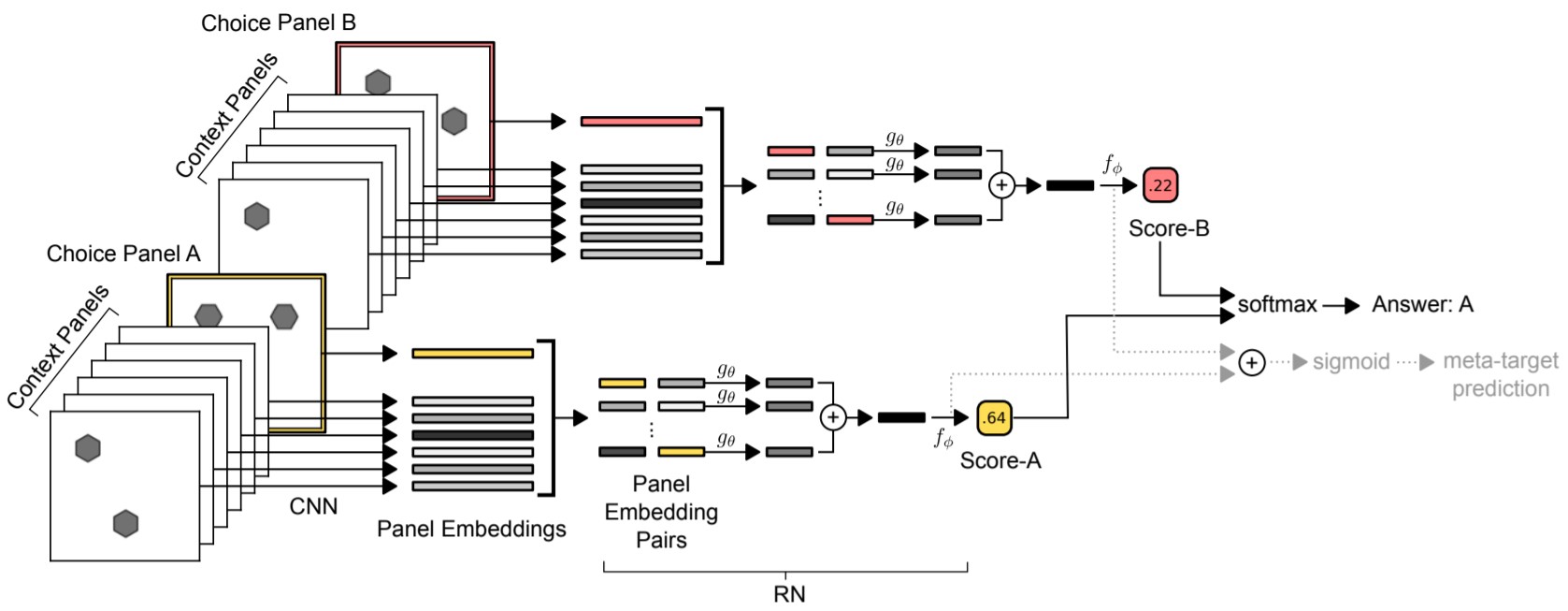}
  \caption{WReN model from \cite{WReN}: A CNN processes each context panel and an individual answer choice panel independently to produce 9 vector embeddings. This set of embeddings is then passed to an RN network \cite{WREN_orig}, whose output is a single sigmoid unit encoding the “score” for the associated answer choice panel. 8 such passes are made through this network (here we only depict 2 for clarity), one for each answer choice, and the scores are put through a softmax function to determine the model’s predicted answer.}
  \label{fig:WReN_architecture}
\end{figure}

To solve a PGM problem, Barrett et al. propose a Wild Relation Network (WReN) architecture~\cite{WREN_orig} as shown in Figure~\ref{fig:WReN_architecture}. In this architecture, all given images (the 8 context panels and the 8 choice panels, all represented as 80x80 grayscale images) are first processed by a small convolutional neural network (CNN), resulting in 16 feature embeddings (one per panel). The 8 context embeddings are then sequentially combined with each option embedding, yielding a total of 8 stacks of 9 embeddings. These are finally processed by the WReN network, yielding a single scalar value for each choice panel, indicating its `matching-score' with the given problem. The entire pipeline is then trained to produce the label of the correct missing panel as an output answer by optimizing a cross entropy loss using stochastic gradient descent. To include spatial information in the panel embeddings, each CNN embedding is also concatenated with a one-hot label indicating the panels position, followed by a linear projection.

Although WReNs achieve reasonable performance ($62.6\%$ classification accuracy) on a randomly held-out test set of the complete training data (which includes triples $S$ from all possible primitives $[r, o, a]$), the generalization performance on new reasoning problems $S$ (containing primitives not seen during training) is significantly worse. This shows that while the model manages to fit the training distribution reasonably well, it fails to generalize in a meaningful way. One of the reasons for this lack of strong generalization is that there is no explicit pressure for the model to discover the generative, latent factors of the problem domain. In fact (as can be seen in Figure \ref{fig:cnn_reconstructions} of the appendix), the learned CNN embeddings seem to completely disregard the underlying causal structure of the problem domain. In this paper, we aim to improve the generalization performance of WReNs, by first learning a disentangled latent space that encodes the PGM panels, and then learning to reason within this space using the RN architecture.

\section{Unsupervised representation learning with Variational Autoencoders}

Because the structure of the Raven problems depend explicitly upon a set of generative factors (such as shape, size, colour, ...), recovering these variables in a suitable latent space should prove beneficial for solving the relational reasoning problem. To test this hypothesis, we leverage Variational Autoencoders \cite{VAE} to learn an unsupervised mapping from high-dimensional pixel space to a lower dimensional and more structured latent space that is subsequently used by the WReN model to complete the relational reasoning task.

The behavior of these models has been widely studied \cite{Disentangling_by_factorising, GECO, Understanding_beta_VAE, Visual_concept_learning_DeepMind} and a clear trade-off between desirable latent space properties (such as disentanglement of generative factors, linearity \& sparsity) and reconstruction quality is usually present. The effect of these constraints on the generalization strength of the resulting latent space, however, has not been widely studied.

As such, our setup replaces the CNN-encoder of \cite{WReN} with a disentangled `$\beta$-VAE' that is trained separately from the WReN model using the modified ELBO optimization objective as described in~\cite{Understanding_beta_VAE}: 
\begin{equation}
{ L } ( \theta , \phi ; \mathbf { x } , \mathbf { z } , \beta ) = \mathbb { E } _ { q _ { \phi } ( \mathbf { z } | \mathbf { x } ) } \left[ \log p _ { \theta } ( \mathbf { x } | \mathbf { z } ) \right] - \beta D _ { K L } \left( q _ { \phi } ( \mathbf { z } | \mathbf { x } ) \| p ( \mathbf { z } ) \right).
\end{equation}
\\In this case, different $\beta$-values control the trade-off between reconstruction quality and latent variable disentanglement.

\begin{figure}[h!]
  \centering
  \includegraphics[width=.95\textwidth]{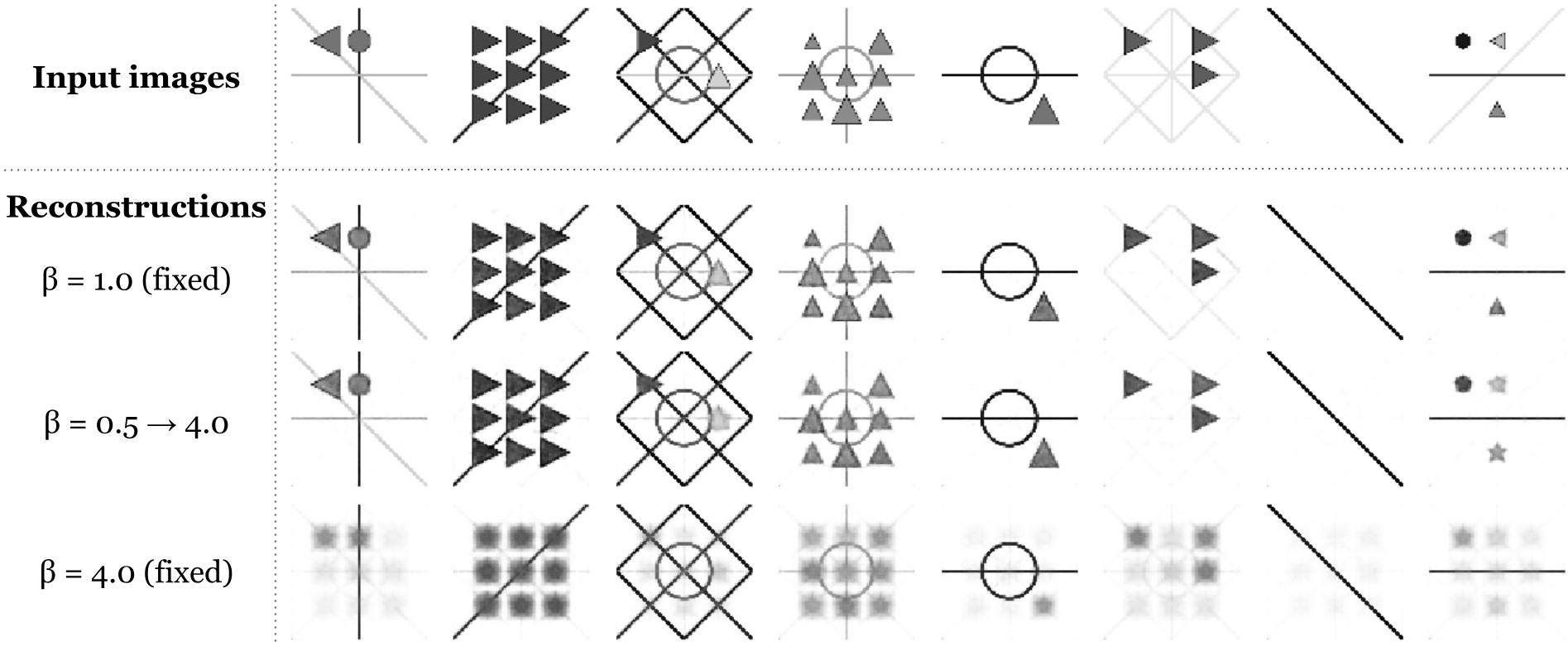}
  \caption{Effect of $\beta$ on reconstruction quality. Three sets of reconstructions are shown using the same VAE trained with different $\beta$ training regimes. In the variable-$\beta$ scenario the latent space first learns to capture small visual details and only later receives pressure to disentangle them.}
  \label{fig:variable_beta_effect}
\end{figure}

One common problem with the $\beta$-VAE setting is that high $\beta$-factors often constrain the latent space to such an extent that encoding small, visual details (which only have a marginal effect on reconstruction error) does not outweigh the KL-penalty that follows from the corresponding divergence from the imposed prior distribution, even though these details are often task-critical. To solve this problem we apply a variable-$\beta$ training regime as described in Appendix \ref{beta_annealing_app}. We ended up using the objective function from $(1)$ with a gradually increasing $\beta$: $0.5 \rightarrow 4.0$. The effects of this training regime can be seen in Figure \ref{fig:variable_beta_effect}.

We trained various models with $z=64$ latent dimensions on the PGM dataset using different $\beta$ factors and annealing schemes. Details of our VAE encoder and decoder architecture and training scheme are discussed in Appendices \ref{VAE_arch_appendix} and \ref{beta_annealing_app}. After training we can visualize that the disentangled latent space indeed captures many of the underlying generative factors of the problem domain (see Figure \ref{fig:beta_vae}). A more extended set of latent space interpolations is shown in Appendix \ref{Latent_space_app}.

\begin{figure}[h!]
  \centering
  \includegraphics[width=.95\textwidth]{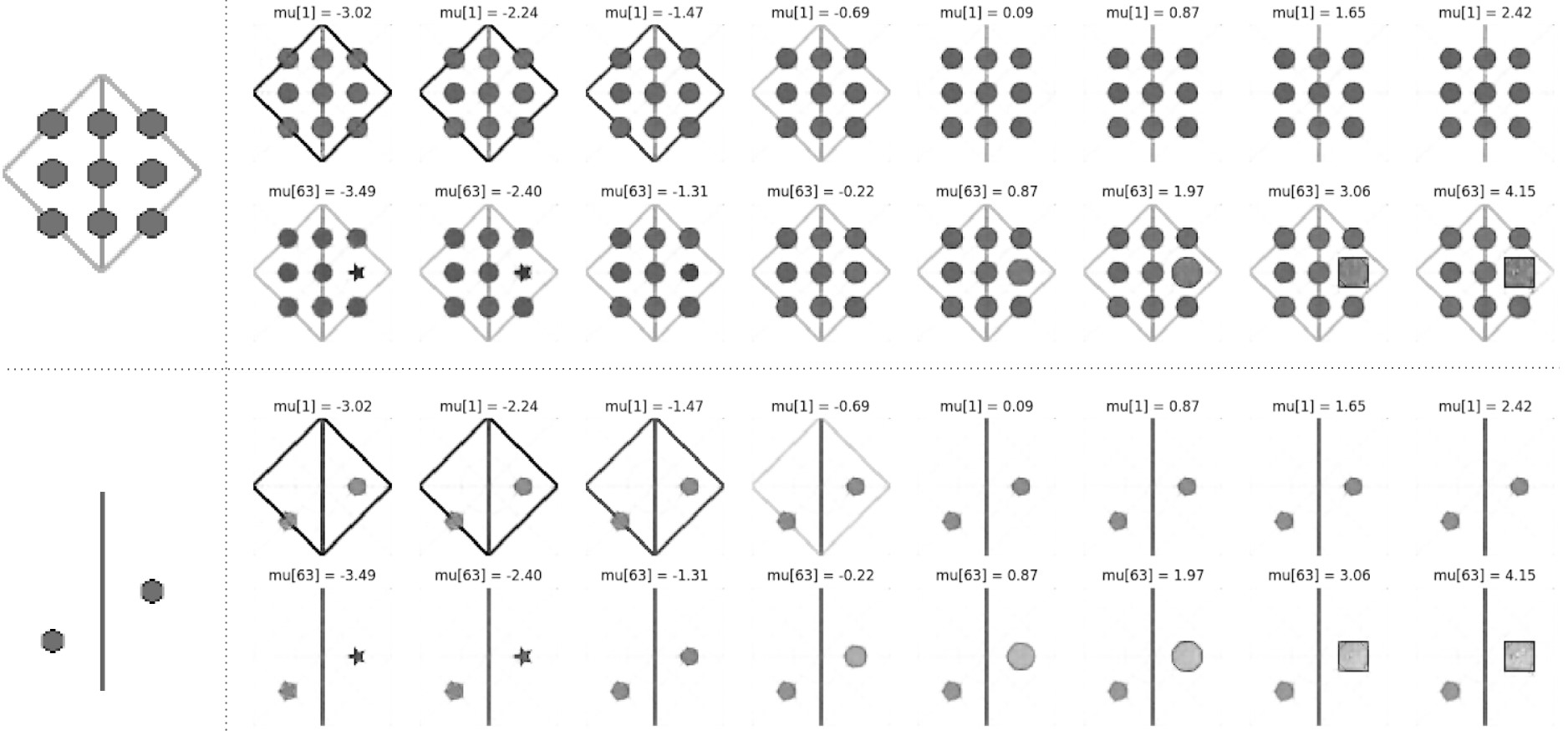}
  \caption{Latent space visualization obtained by encoding the input images (left) and interpolating between the support boundaries of the posterior distribution for $z[1]$ and $z[63]$, while keeping all other latents constant before rendering the resulting $z$ through the decoder network. Notice how the VAE's latent space ($\beta$=4.00) clearly disentangles multiple generative factors such as the colour/presence of the diamond-shaped raster (top rows) or the shape of an object in a single position (bottom rows).}
  \label{fig:beta_vae}
\end{figure}

\section{Leveraging the learned latent space for relational reasoning}

Here we test the generalization properties of the learned latent space by using the learned image embeddings in the PGM problem setting. 
To start the WReN training process, we freeze the pretrained VAE-encoder graph and use it to initialize the encoder step of the WReN architecture. In order to create a fair comparison, our VAE architecture uses the exact same encoder network as the CNN-embedder in \cite{WReN}. But, since we use a latent space of $z=64$ latent dimensions, the $512$ convolutional-features from the encoder are passed through two FC-layers mapping onto $64$-dimensional vectors representing the means and $log($variances$)$ of a factorized, Gaussian distribution. At training time we randomly sample from this posterior to get the latent representations used in further processing. At test time, we simply use the mean vector. Apart from this difference in input feature dimensionality (64 in our VAE case vs 512 in the default CNN-encoder case), the entire WReN architecture is identical to the one used in \cite{WReN}.

Finally, the WReN model is trained for 6 epochs using a fixed encoder and then finetuned end-to-end for another 2 epochs to get the results displayed in Table \ref{table:WReN}. Notice that while we outperform the default WReN model trained with purely supervised learning on various generalization regimes as intended, surprisingly we also do better on two of the validation sets, indicating that the disentangled latent space does in fact make the relational reasoning problem more tractable for the RN network.

\begin{table}[h!]
  \label{sample-table}
  \centering
  \begin{tabular}{lllllll}
    \toprule   
    \cmidrule(r){1-7}
    \multicolumn{1}{c}{Model-type} & \multicolumn{3}{c}{CNN-WReN \cite{WReN}} & \multicolumn{3}{c}{VAE-WReN ($\beta$=4.00)} \\
    \cmidrule(r){1-7}
    Generalization regime& Val (\si{\percent})& Test (\si{\percent})& Test (kappa)& Val (\si{\percent})& Test (\si{\percent})& Test (kappa)\\
    \midrule
    Neutral & 63.0& 62.6 &0.573 & \textbf{64.8} & \textbf{64.2} & \textbf{0.591}\\
    H.O. Triple Pairs& 63.9& 41.9 &0.336 &\textbf{64.6} & \textbf{43.6} &\textbf{0.355}\\
    H.O. Attribute Pairs& 46.7& 27.2&0.168 &\textbf{70.1} &\textbf{36.8} &\textbf{0.278}\\
    H.O. Triples& \textbf{63.4}& 19.0 &0.074 &59.5 &\textbf{24.6} &\textbf{0.138}\\
    \bottomrule
  \end{tabular}
  \caption{Relational Reasoning Results. Each data regime consists of 1.2M training images, a held-out validation set of 20K images drawn from the same problem distribution and a generalization test set of 200K images containing new problem sets $S = \{r, o, a\}$ not seen during training. Regimes are sorted according to the degree of generalization required to solve them (more info in \cite{WReN}). We also list Cohen's Kappa values which vary linearly from $0$ (random guessing) to $1$ (oracle).}
  \label{table:WReN}
\end{table}

\section{Conclusion}
In this paper, we show that disentangled variational autoencoders can be leveraged to learn a mapping from high-dimensional pixel space to a low-dimensional and more structured latent space without any explicit supervision. This disentangled latent space can subsequently be leveraged for solving a non-trivial relational reasoning problem and by doing so, outperforms the same architecture trained using a fully supervised approach.

\subsection{Future work}
Future work will focus on further investigating the desirable characteristics of a generally useful latent space (disentanglement, linearity, sparsity, ...) and explore new objective functions that can be leveraged for unsupervised representation learning, such as various representation losses \cite{DARLA, beyond_pixels}, GAN-inspired discriminator networks \cite{CVAEGAN} and predictive capacity \cite{ConsciousnessPriorBengio, ContrastivePredictiveCoding}. Additionally, the current WReN setup does not leverage the intrinsic variational properties of the learned latent space. 


\bibliographystyle{unsrt}
\bibliography{bibliography}

\section*{Appendix}
\appendix
\section{Architecture details} \label{VAE_arch_appendix}
Our VAE setup uses a convolutional encoder-decoder architecture and a $N(0,1)$-Gaussian distribution as prior for the latent variables. The encoder has 4 $conv2d()$ layers intermitted with $BatchNorm2d()$ layers. The decoder uses the same architecture with $ConvTranspose2d()$ layers. Every conv-layer uses 32 kernels of size 3 and a stride of 2. We did not use any pooling layers.

The VAE bottleneck has 64 latent variables which are parameterized through their respective means and $log($variances$)$. We use the common reparameterization trick from \cite{VAE} for backpropagating through the non-deterministic bottleneck.

The VAE was trained using the ADAM optimizer with a learning rate of .0003 and a batch size of 32 PGM problems per batch ( = 512 panel images).

In the WReN network, we use dropout $(p=0.5)$ in the penultimate layer.
All models were implemented in PyTorch and trained on a single Tesla K80 GPU.

\section{Additional latent space visualizations}
\label{Latent_space_app}
\begin{figure}[h!]
  \centering
  \includegraphics[width=\textwidth]{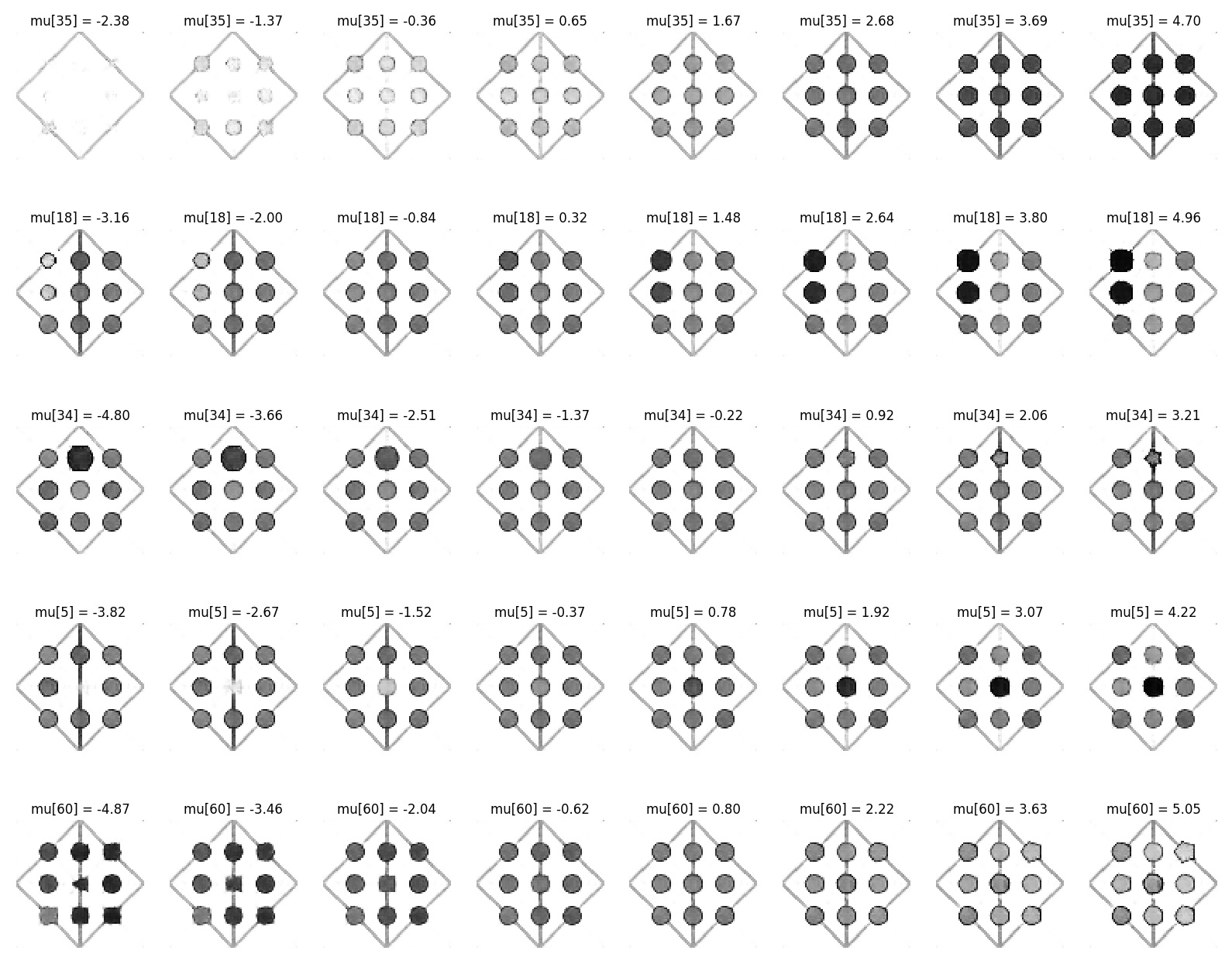}
  \caption{Latent space interpolations. VAE with $\beta$=0,01 - avg-MSE = 6,06 and avg-KL-divergence = 120. Many latent traversals change a variety of generative factors simultaneously. Note that the numerical bounds of the interpolations for each latent variable are clipped using the support from the posterior distribution (generated by passing 5000 random train images through the encoder network).}
  \label{fig:low_beta_interpolations}
\end{figure}

\begin{figure}[h!]
  \centering
  \includegraphics[width=\textwidth]{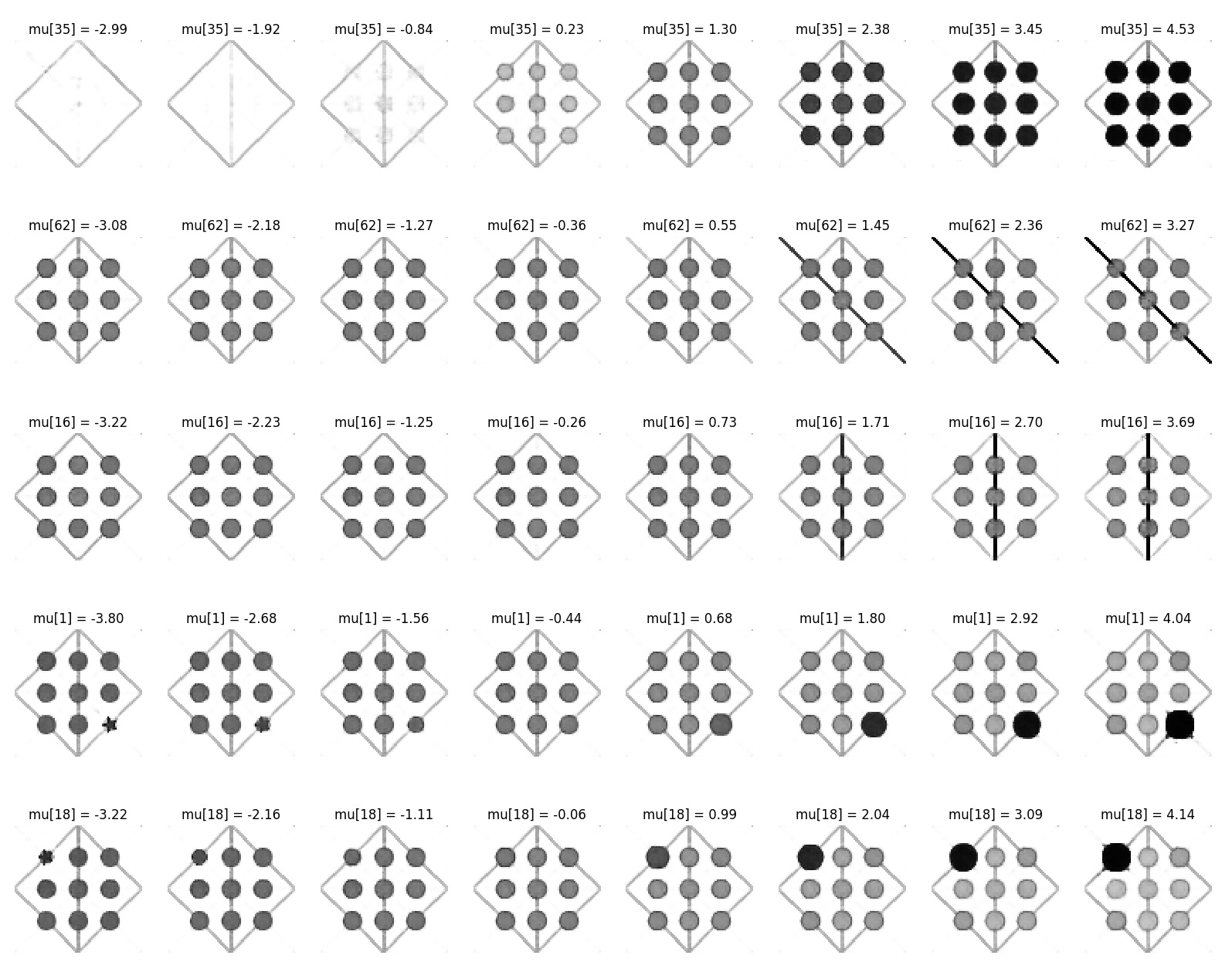}
  \caption{Latent space interpolations. VAE with $\beta$=4,00 - avg-MSE=20,2 and avg-KL-divergence=22. Most latent traversals correspond to a single, clear generative factor. Again, the interpolation bounds are clipped using the support from the posterior distribution.}
  \label{fig:high_beta_interpolations}
\end{figure}

\begin{figure}[h!]
  \centering
  \includegraphics[width=\textwidth]{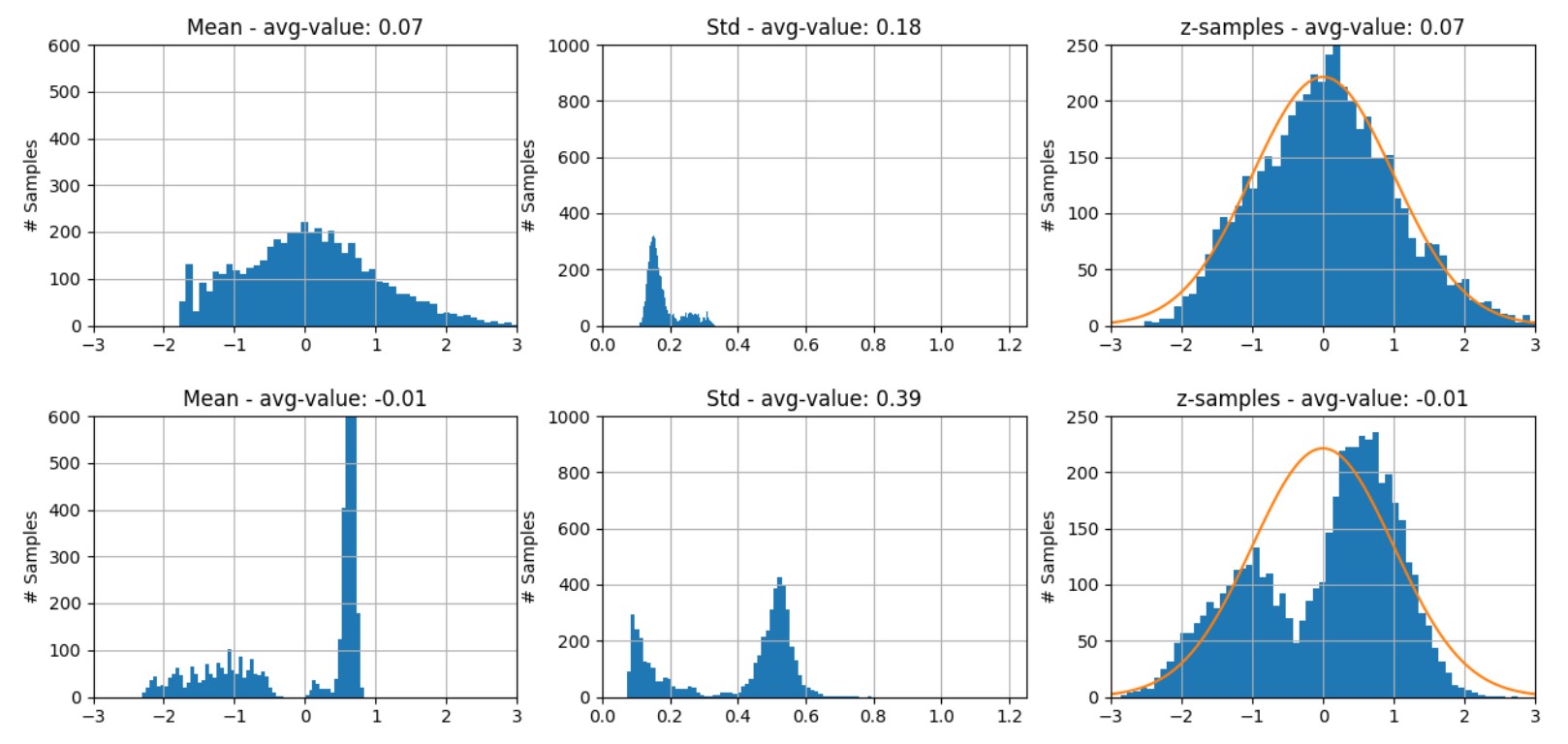}
  \caption{Visualization of the latent distribution for the two variables with highest average KL-divergence from a disentangled VAE with $\beta$=4.0. We run 5000 randomly sampled images through the encoder network and plot the resulting distributions of $\mu$ and $\sigma$ as well as 5000 random samples drawn from each of the resulting Gaussian distributions. Notice that latent variables with a very high KL-divergence from the $N(0,1)$-prior (eg. top row where $\sigma$ is always close to zero) can still result in a near-Gaussian distribution over sampled z-values.}
  \label{fig:latent_distributions}
\end{figure}

To further aid the insight of the reader, we provide a bunch of additional visualizations we found helpful to understand the latent space behavior of disentangled VAEs.
See Figures \ref{fig:low_beta_interpolations}, \ref{fig:high_beta_interpolations}, \ref{fig:latent_distributions}.

\section{Failure modes of the WReN CNN} \label{CNN_failure_app}
\begin{figure}[h!]
  \centering
  \includegraphics[width=0.8\textwidth]{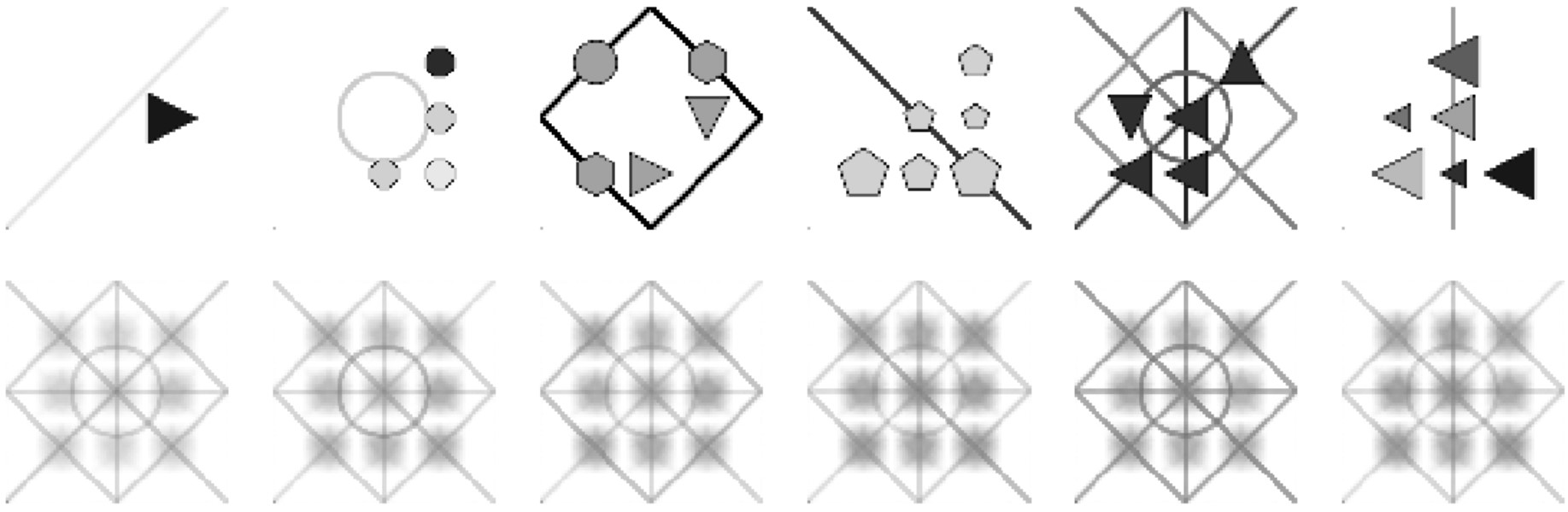}
  \caption{Input images (top) and reconstructions (bottom) obtained by training a convolutional-decoder on the panel features extracted by the CNN embedder from \cite{WReN}. As can be seen, the generative factors defining the problem domain are not contained within the learned embedding space, lending support to the claim that the CNN simply overfits on specific visual features in the training examples instead of discovering useful latent structure.}
  \label{fig:cnn_reconstructions}
\end{figure}

One of the problems with the purely supervised CNN approach is that the model receives no explicit pressure to discover the generative, latent structure of the problem domain. This can be clearly seen in Figure \ref{fig:cnn_reconstructions}.

\begin{figure}[h!]
  \centering
  \includegraphics[width=.8\textwidth]{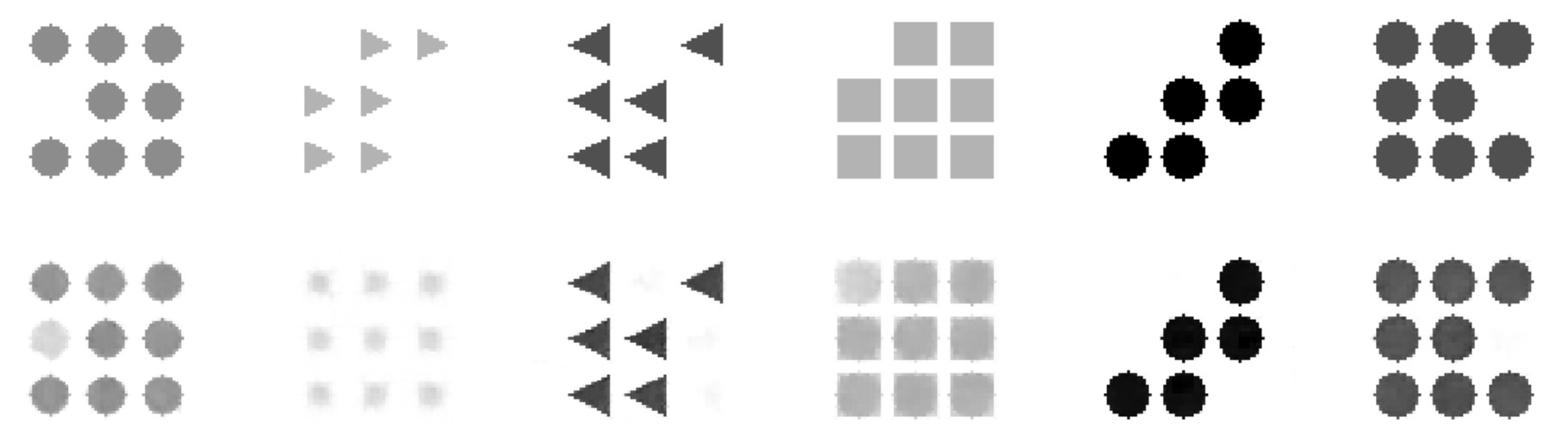}
  \caption{When forcing the VAE to trade-off reconstruction quality for smaller KL-divergence from the latent prior, small, grayscale objects will be sacrificed first since they correspond to the smallest increase in MSE penalty in pixel space. Larger and darker objects will maintain good reconstruction quality until much higher $\beta$ values are imposed. (Shown here on a custom dataset we generated for testing purposes.)}
  \label{fig:gray_loss}
\end{figure}

Unfortunately, the used VAE approach based on a pixel-reconstruction loss also has its own drawbacks as can be seen in Figure \ref{fig:gray_loss}. This lends support to the widely held assumption (see eg. \cite{DARLA, beyond_pixels}) that simple, pixel-based reconstruction metrics are not the ideal optimization objectives for visual representation learning.

\section{Impact of annealing \texorpdfstring{$\beta$}{Lg}}\label{beta_annealing_app}

One common problem with $\beta$-VAEs is that imposing a large disentanglement constraint often causes the latent space to collapse to supporting only the most salient modes of the visual input domain, failing to capture more fine-grained visual information that is often task critical.

To tackle this problem we start training the VAE with a low $\beta$-factor and gradually increase the disentanglement constraint until a desirable state is reached. By primarily focusing on the visual reconstruction error in the beginning of training, the latent space learns to capture most of the visual information before the disentanglement constraint begins dominating the training objective, leading to a much better final representation model. 

\end{document}